# Scribble-based Weakly Supervised Deep Learning for Road Surface Extraction from Remote Sensing Images

Yao Wei, and Shunping Ji, *Member*, *IEEE*

*Abstract*—Road surface extraction from remote sensing images using deep learning methods has achieved good performance, while most of the existing methods are based on fully supervised learning, which requires a large amount of training data with laborious per-pixel annotation. In this paper, we propose a scribble-based weakly supervised road surface extraction method named ScRoadExtractor, which learns from easily accessible scribbles such as centerlines instead of densely annotated road surface ground-truths. To propagate semantic information from sparse scribbles to unlabeled pixels, we introduce a road label propagation algorithm which considers both the buffer-based properties of road networks and the color and spatial information of super-pixels. The proposal masks generated from the road label propagation algorithm are utilized to train a dual-branch encoder-decoder network we designed, which consists of a semantic segmentation branch and an auxiliary boundary detection branch. We perform experiments on three diverse road datasets that are comprised of high-resolution remote sensing satellite and aerial images across the world. The results demonstrate that ScRoadExtractor exceed the classic scribble-supervised segmentation method by 20% for the intersection over union (IoU) indicator and outperform the state-of-the-art scribble-based weakly supervised methods at least 4%.

## I. Introduction

As a fundamental and important problem in the field of remote sensing image processing, road extraction has a great number of applications including navigation, geo-information database updating, disaster management, and autonomous driving; and also provides contextual information that benefits other related tasks, such as land cover classification and vehicle detection. In general, road extraction from remote sensing imagery falls into two subtasks: road surface extraction [1, 2, 3, 4, 5] and road centerline extraction [6, 7, 8, 9, 10]. The former focuses on extracting the complete road surfaces from backgrounds as a binary segmentation map, whereas the latter aims to extract the topological structure of road networks in vector form without road width information. With the rapid development of volunteered geographic information (VGI) sources which allow millions of contributors to create and edit geographic data across the world, more road vector data are becoming publicly available. However, these open-source maps are often noisy and/or incomplete. For example, OpenStreetMap (OSM) [11], which is one of the most extensive VGI sources, provides only the coordinates of road centerlines without road width information. Therefore, new effective methods are expected to achieve accurate and automatic road surface extraction from remote sensing images and the existing OSM centerlines with minimum human cost.

Recently, the deep learning (especially deep convolutional neural network (DCNN)) based road surface extraction methods have been widely studied and achieved good performance, however, these methods are based on fully supervised learning, i.e., a huge amount of per-pixel annotation of road surfaces has to be prepared as training samples, which demands laborious work and is hardly met in practice. To avoid such demand, weakly supervised learning, which attempts to learn from low-cost sparse annotation (e.g., scribble, click), has drawn increasingly more attention in computer vision fields. This motivates the road surface extraction to be formulated as a weakly supervised deep learning task.

In this paper, we investigate the possibility of training a weakly supervised deep learning model with existing and easily accessible scribbles such as road centerlines derived from manual delineation, OSM data or GPS traces. We attempt to extract road surface through learning from the scribble annotations, making it possible to greatly reduce the annotation effort. Apparently, this consideration is highly related to the development of VGI sources as noted above and the lack of perfect supervision in real-world applications.

The main contributions of our research are as follows. 1) We propose a novel scribble-based weakly supervised deep learning approach (called ScRoadExtractor) for road surface extraction from remote sensing images under the weak supervision of centerline-like scribble annotations. 2) A road label propagation algorithm is proposed to propagate the semantic information from scribbles to unlabeled pixels by utilizing both the buffer-based properties of roads and the local

Manuscript received August 17, 2020; revised October 24, 2020. This work was supported by the National Key Research and Development Program of China under Grant 2018YFB0505003. (Corresponding author: Shunping Ji).

Y. Wei, and S. Ji are with the School of Remote Sensing and Information Engineering, Wuhan University, Wuhan, HB 430079, China (e-mail: weiyao@whu.edu.cn; jishunping@whu.edu.cn). The code is available at: *https://github.com/weiyao1996/ScRoadExtractor*.

and global dependencies between graph nodes built on super-pixels to generate the proposal mask. 3) We design a dual-branch encoder-decoder network (DBNet), which is trained with the proposal mask and boundary prior information detected from images, and outputs a road surface segmentation map that approaches to a map predicted from a densely-supervised method. 4) The experimental results on diverse road datasets across the world demonstrate that our approach possesses high-performance and powerful generalization ability and also outperform state-of-the-art scribble-supervised segmentation methods.

The remainder of this paper is arranged as follows. In Section II, we briefly review the related studies. Section III provides a detailed description of ScRoadExtractor, including the road label propagation algorithm and the dual-branch encoder-decoder network. Section IV presents the experiments we conducted to verify the effectiveness and generalization ability of the proposed method on diverse datasets with comparison to most recent studies. Discussions are given in Section V, and Section VI presents our conclusions and future research prospects.

## II. RELATED WORK

We briefly review the fully supervised deep learning methods for road surface extraction, the generic scribble-based weakly supervised methods, and a few weakly supervised learning methods designed for road surface extraction.

Mnih and Hinton [1] adopted a deep belief network composed of restricted Boltzmann machines (RBMs) for road surface extraction. Panboonyuen et al. [12] employed SegNet [13], a variant of fully convolutional network (FCN), in road segmentation and implemented post-processing using landscape metrics and conditional random field (CRF) [14]. Zhang et al. [15] proposed an improved DCNN, which combined ResNet [16] and U-Net [17] to extract roads from remote sensing images. Xu et al. [18] utilized a global and local attention model based on U-Net and DenseNet [19]. Zhou et al. [20] proposed D-LinkNet model for the road segmentation task, which combined the benefits of encoder-decoder architecture and dilated convolution to capture multi-scale features. He et al. [21] improved the performance of road extraction networks by integrating the Atrous spatial pyramid pool (ASPP) [22] with an encoder-decoder network to enhance the ability of extracting the detailed features of the road. Wei et al. [23] proposed a DCNN-based framework that aggregated the semantic and topological information of road networks to produce refined road surface segmentation maps with better connectivity and completeness.

However, training such networks relies on large amounts of densely annotated labels for optimizing millions of parameters. The data-driven supervised learning approaches may be not practical for real-world applications due to the lack of perfect supervision and the demand of labor-intensive and time-consuming works.

In order to tackle this issue, weakly supervised learning has been explored to avoid annotating huge amounts of training data, which involves learning from weak supervision such as scribble [24, 25], click [26], bounding box [27, 28] and image-level tags [29, 30]. Typically, most of the existing weakly supervised learning methods adopt an alternative training scheme: generate pseudo-semantic labels (i.e., proposals) from the seeds provided by sparse annotated data; train deep models (e.g., DCNNs) with these proposals using standard loss functions (e.g., cross-entropy); alternate between the proposal generation and the network training steps. Based on scribble annotations and graph theory [31, 32], Lin et al. [24] alternated between generating proposals using a graph defined over super-pixels and training an FCN with the proposals, which grew more reliable along with the energy function of the graphical model was optimized with FCN predictions iteratively. However, the method assumed that the labels were constant within super-pixels, which led to an artificial upper bound on the accuracy of proposal generation. Similarly, Papandreou et al. [33] alternated between two steps in an iterative manner: estimating the latent pixel labels through improved expectation-maximization methods from bounding box annotations and image-level annotations, and training DCNNs in weakly-supervised and semi-supervised settings.

Although the quality of the generated proposals can be enhanced by alternating optimization, these approaches share a common drawback that the training tends to be vulnerable to inaccurate intermediate proposals which are treated as labels, since standard loss functions do not distinguish the seeds from the mislabeled pixels.

Hence, additional regularizations including graph-based approaches (e.g. CRF) and boundary-based losses [34] are often employed to enrich the semantic information of weak annotations in an end-to-end manner. Tang et al. [35, 36] introduced a normalized cut loss and a partial cross-entropy loss for weakly supervised semantic segmentation, and incorporated standard regularization techniques (graph cuts and dense CRFs) into the loss function over the partial inputs. Obukhov et al. [37] proposed a gated CRF loss for unlabeled pixels together with partial cross-entropy loss for labeled pixels. Lampert et al. [34] applied a constrain-to-boundary principle to recover detail information for weakly supervised segmentation. More recently, boundaries have been directly embedded into segmentation network. Wang et al. [38] designed a network architecture that consisted of two sub-networks: the prediction refinement network (PRN) and the boundary regression network (BRN), where the BRN guided the PRN in localizing the boundaries. However, the semantics and boundaries information interact only at the loss functions, without considering the correlation of the features between two sub-networks. Zhang et al. [39] proposed a weakly-supervised salient object detection method by introducing an auxiliary edge detection network and a gated structure-aware loss which focused on the salient regions of images. However, this method was not very stable due to its over-confident saliency predictions.

The aforementioned weakly supervised learning methods had yet to be recognized as promising for road surface extraction until recently. Several studies [40, 41] attempted to employ publicly available maps (e.g., OSM vector data) for road surface extraction from aerial images; however, these methods did not utilize the recent weakly supervised deep learning techniques as mentioned above, and only considered the VGI-based road segmentation as a width estimation problem, making it not robust to the occlusions.

In terms of road vector data, pixels on the road vectors can provide weak supervision for segmenting road surface, which motivates road extraction being formulated as a weakly supervised learning task by two most recent studies. Kaiser *et al.* [42] trained a CNN for building and road extraction with noisy labels generated from OSM data; specifically, the road labels were simply determined by an average road width for each category (e.g., highway, motorway) which was provided by OSM. According to a predefined road width, Wu *et al.* [43] utilized OSM centerline to produce initial road annotation masks which were then fed into a road segmentation network using the partial loss for labeled pixels and the normalized cut loss for unlabeled pixels. It is obvious that the main shortcoming of these two methods is the fixed width supervision, which is a simple and empirical design strategy, only deals with specific data and limits its application for large-scale data sets. Thus, there are still many more aspects that should be investigated in more detail for weakly supervised deep learning-based road surface extraction.

## III. METHODOLOGY

We propose ScRoadExtractor, a novel scribble-based weakly supervised deep learning method for road surface extraction from remote sensing images, the framework is illustrated in Fig. 1. First, the proposal masks are generated by a road label propagation algorithm based on the remote sensing images and scribble annotations. Then, the generated proposal masks and boundary prior information detected from images are used to train a dual-branch encoder-decoder network for road surface extraction by minimizing a joint loss function.

### A. Road Label Propagation

Since scribble annotations provide sparse information that limits the overall accuracy of labeling, directly training a DCNN model with sparse scribbles inevitably leads to poor identification results. When taking road centerlines derived from GPS traces or OSM as scribbles, a straightforward method is to expand the centerline with a predefined width, but it cannot perfectly identify road boundaries as road width varies. Another possible solution is to mark pixels with features similar to road pixels as roads. Starting from the two straight ideas, we develop a more sophisticated context-aware road label propagation algorithm which propagates the semantic labels from scribbles to unlabeled pixels and marks every pixel of an image within two categories: known (road and non-road) and unknown pixels, as shown in Fig. 2.

First, considering the property of road networks that the boundaries tend to be parallel to the road centerline, a buffer-based strategy is applied to infer buffer-based masks according to the distance from the road centerline. Specifically, two buffers of scribbles are created with buffer width $a_1$ and $a_2$ ($a_1 < a_2$) respectively. The pixels within the first buffer are denoted as the road pixels, the pixels outside the second buffer represent the non-road pixels, and the remaining pixels are categorized as unknown pixels. However, the buffer-based strategy only generates coarse masks and relies on the quality of the scribbles, for example, incorrect and incomplete centerlines may exist in outdated GIS maps or OSM data.

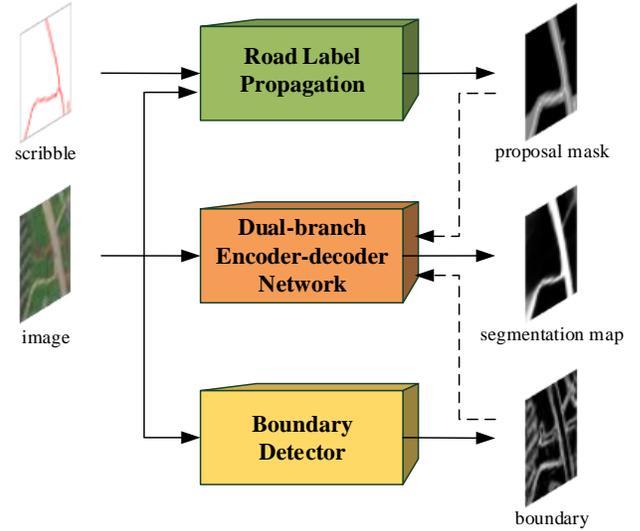

Fig. 1. Overview of the proposed ScRoadExtractor in training phase. The dash line represents the constraints in the loss function.

Second, a graphical model is constructed on the super-pixels of a training image by minimizing an energy function, which was inspired by Graph Cut [31, 32] that leverages the unary and pairwise potentials to model the local and global dependencies between graph nodes. First, we employ the simple linear iterative clustering (SLIC) [44] to generate the super-pixels. Second, we convert the images from red, green, blue (RGB) space to the hue saturation value (HSV) space, and then the color histograms for all the super-pixels, which are 2-dimensional over the H and S channels, are calculated. The super-pixels that overlap with the scribbles are adopted as the foreground (road) samples, and the super-pixels that overlap outside the $a_2$ buffer are designated as background (non-road) samples. Accordingly, the cumulative histograms for the foreground and background are calculated. Third, a graph is built where a node represents a super-pixel. For each node, there are two types of corresponding edges. A type of edges connects the node with its neighbor nodes, and the other type of edges connects it with both foreground and background nodes. The energy function is defined below:

$$E(x) = \sum_i \psi_i(x_i | Hist, Sc) + \sum_{i,j} \psi_{ij}(x_i, x_j | Hist) \quad (1)$$

*Hist* denotes the color histograms for all super-pixels, $Sc = \{s_r, c_r\}$ denotes the scribble annotations where $s_r$ is the pixels of scribble $r$ and $c_r \in \{foreground, background\}$ is the category label of scribble $r$. The unary term $\psi_i$ can be formulated as follows:

$$\psi_i(x_i) = \begin{cases} 0 & \text{if } x_i \cap s_r \neq \varnothing \text{ and } c_i = c_r \\ \text{KLDiv}(Hist_i, Hist_r) & \text{if } x_i \cap Sc = \varnothing \\ \infty & \text{otherwise} \end{cases} \quad (2)$$

In the first condition, if super-pixel $x_i$ overlaps with $s_r$, then it has zero cost when being assigned the label $c_r$. The second condition means that if $x_i$ does not overlap with any scribble, then the cost is calculated through Kullback-Leibler divergence (KLDiv) between the normalized histogram of $x_i$ and the cumulative histogram of $c_r$.

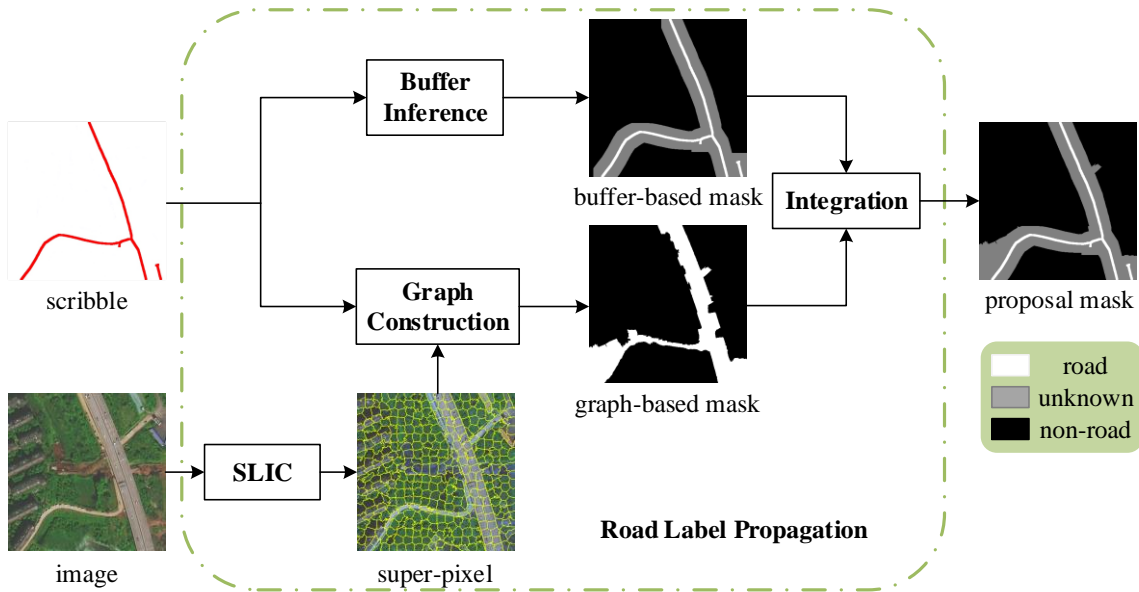

Fig. 2. Process of the road label propagation algorithm.

The pairwise term $\psi_{ij}$ evaluates the appearance similarities between two neighbor super-pixels (i.e., $x_i$ and $x_j$ ($x_i \neq x_j$)), by comparing their normalized color histograms using KLDiv. It is assumed that the cost of cutting the edge between two neighbor super-pixels with closer appearance is higher, namely, two neighbor nodes with larger similarities are more likely to have the same label. The graph model propagates label information from the scribbles to the unlabeled pixels to generate a road mask map, but the graph-based masks may contain errors due to the highly varying appearances of the images and the limited capacity of the graph cut based method.

Finally, we make full use of the buffer inference and graph construction. We integrate the buffer-based and graph-based masks based on the following cues: if the pixels denote road in the graph-based mask and non-road in the buffer-based mask, they are marked as unknown pixels, and the remaining pixels are assigned the same as the buffer-based mask. In this way, we obtain proposal masks which consider not only the buffer-based attributes of road networks but also the color and spatial information obtained from the graph constructed on the super-pixels of the training images. The advantage of our context-aware label propagation algorithm is explicit. The mislabeled pixels from the graph-based masks are hardly distinguished by a standard loss function, which would inevitably impact the results of segmentation; the buffer-based masks assert the absolute discrimination of road and non-road. In contrast, the labels of unknown pixels of our proposal masks are changeable in a learning-based scheme. The unknown pixels of the proposal masks are classified into potential road or non-road pixels iteratively through the regularized weakly supervised loss that is described in the next section.

### B. Dual-branch Encoder-decoder Network

As illustrated in Fig. 3, the architecture of our dual-branch encoder-decoder network (DBNet) consists of three parts (one encoder and two decoders) with the RGB channels of an image at a size of $512 \times 512$ as input. The first part is a feature encoding network using ResNet-34 [16] pretrained on ImageNet [45], and it has five downsampling layers with a minimum scaling ratio of 1/32. Except for the $7 \times 7$ convolution layer with stride 2 and the first residual blocks with channel number 64, the feature channels are doubled at each downsampling step in the encoder. With regard to the decoding stage, we design two sub-network branches in parallel: the semantic segmentation branch and the boundary detection branch. Specifically, the segmentation branch uses five transposed convolution layers [46] with stride 2 to restore the resolution of the feature maps from $16 \times 16$ to $512 \times 512$, and the feature channels are halved at each upsampling step except for the last two layers. There are three addition skip connections (denoted as circled +) between the encoder feature maps and the decoder (segmentation branch) feature maps. The feature maps with size $256 \times 256$ in the segmentation branch are concatenated with the corresponding feature maps in the boundary branch. The ASPP module [22], which consists of (a) one $1 \times 1$ convolution and three parallel $3 \times 3$ Atrous convolutions with Atrous rates of 1, 2, and 4, respectively, and (b) global average pooling, is applied to the last feature maps of the encoder. The resulting feature maps of ASPP are concatenated and passed through the $1 \times 1$ convolution layer with channel number 512, and finally fed into the decoder (segmentation branch) part.

For the boundary branch, the multi-scale features extracted from the segmentation network are reused. As denoted in Fig. 3, the 512-dimensional features in the decoder (segmentation branch) are first bilinearly upsampled by a factor of 4 and processed by a $3 \times 3$ convolution layer with channel number 128 and concatenated with the corresponding low-level 128-

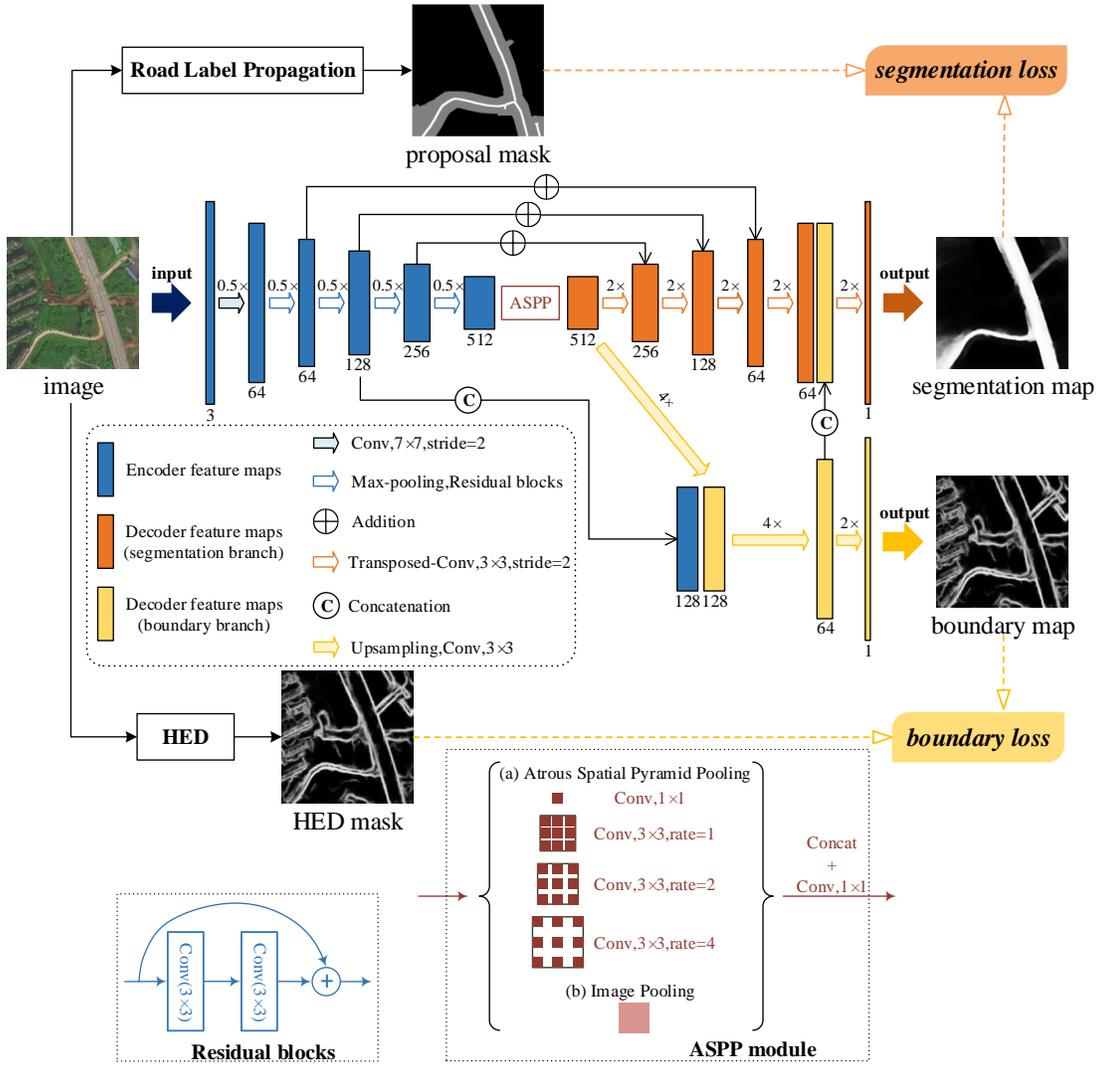

Fig. 3. Structure of the dual-branch encoder-decoder network.

dimensional features from the encoder, followed by another bilinear upsampling by a factor of 4 and a 3 × 3 convolution layer with channel number 64. The feature maps then are bilinearly upsampled by a factor of 2.

Each convolution layer is activated by the rectified linear unit (ReLU) function except the last convolution layer of the two branches which use sigmoid activation to separately output the probability of each pixel belongs to road surface and road boundary, respectively.

DBNet takes the advantages of the encoder-decoder architecture, skip connections, and dual-branch where the boundaries detected by the boundary branch are utilized as prior knowledge to refine and guide the segmentation branch. Here, a holistically-nested edge detection (HED) [47] boundary detector pretrained on the generic boundaries of BSDS500 [48] is applied to produce coarse boundary masks, which we called HED masks, without any fine-tuning. Apart from the HED masks, the proposed network is trained with the proposal masks generated by the road label propagation algorithm described in Section III-A. The segmentation branch and the boundary branch are incorporated under the constraint of a joint loss which combines segmentation loss and boundary loss.

Based on the two categories of labels (known and unknown) provided by the proposal mask, the segmentation loss function is

$$L_{seg} = PBCE(Y_p, S_p) + \alpha R(S) \quad (3)$$

where $PBCE(Y_p, S_p)$ is the partial binary cross-entropy loss described in Equation (4) and only computes the binary cross-entropy loss between proposal mask $Y \in \{0,1\}$ and segmentation map $S \in [0,1]$ for known pixels $p \in \Omega_k$, $R(S)$ denotes the regularized loss [36], which is implemented by the CRF loss with dense Gaussian kernel $W$ over RGBXY channels using fast bilateral filtering [49], and $\alpha$ balances between the two losses.

$$PBCE(Y_p, S_p) = -\sum_{p \in \Omega_k} Y_p \log S_p + (1-Y_p)\log(1-S_p) \quad (4)$$

$$R(S) = S'W(1-S) \quad (5)$$

The gradient of the regularized loss with respect to $S$ is,

$$\frac{\partial R(S)}{\partial S} = -2WS \quad (6)$$

The loss function of the boundary branch is defined between HED mask $T$ and boundary map $B$ based on the per-pixel mean squared error (MSE) loss, that is,

$$L_{bound} = \frac{1}{w \times h} \sum_{i=1}^{w} \sum_{j=1}^{h} (T_{ij} - B_{ij})^2 \quad (7)$$

where $w$ and $h$ are the width and height of the boundary map. Overall, the joint loss is

$$L = L_{seg} + \beta L_{bound} \quad (8)$$

where $\beta$ is a coefficient to balance the segmentation loss and the boundary loss.

## IV. EXPERIMENT AND ANALYSIS

### A. Datasets and Evaluation Metrics

We performed our experiments on three diverse road datasets: 1) the Cheng dataset [50]; 2) the Wuhan dataset; and 3) the DeepGlobe dataset [51]. These datasets are comprised of high-resolution aerial and satellite images from urban, suburban, and rural regions covering a total area of approximately 1665 km² across the world with varied ground sampling distance (GSD) from 0.5 m to 1.2 m. All the images and the corresponding ground-truths were seamlessly cropped into 512 ×512 tiles, and then divided into training set and test set. The details of the datasets are listed in Table I, and examples are shown in Fig. 4. In our experiments, the road centerline was utilized as a typical scribble supervision. Specifically, the ground-truth of the Cheng dataset included manually labeled road surface and centerline, whereas there was only pixel-wise annotated road surface ground-truth for the Wuhan dataset and the DeepGlobe dataset; therefore, we skeletonized the road surface ground-truth to obtain the centerline ground-truth for the last two road datasets.

The precision, recall, F₁ score, and intersection over union (IoU), which were widely used as indicators in the related literature on road segmentation (e.g., [12, 20, 21]), were adopted to evaluate the segmentation accuracy of the road extraction results at the pixel level. The precision was the fraction of the predicted road pixels that were true roads, and the recall was the fraction of the true road pixels that were correctly predicted. The $F_1$ and IoU were the overall metrics that offered a tradeoff between precision and recall. More precisely, the IoU was the ratio between the intersection of the predicted road pixels and the labeled road pixels and the results of their union. These pixel-level evaluation metrics are defined as follows:

$$\text{Precision} = \frac{TP}{TP+FP} \quad (9)$$

$$\text{Recall} = \frac{TP}{TP+FN} \quad (10)$$

$$F_1 = \frac{2TP}{2TP+FP+FN} \quad (11)$$

$$\text{IoU} = \frac{TP}{TP+FP+FN} \quad (12)$$

where TP, FP, and FN represent the true positive, false positive, and false negative, respectively.

TABLE I
DETAILS OF THE EXPERIMENTAL ROAD DATASETS

| Dataset | Area (km²) | Source | GSD(m) | Tiles(train/test) |
|---|---|---|---|---|
| Cheng [50] | 132 | aerial | 1.2 | 300 / 49 |
| Wuhan | 200 | satellite | 0.5 | 1944 / 648 |
| DeepGlobe [51] | 1333 | satellite | 0.5 | 15000 / 5333 |

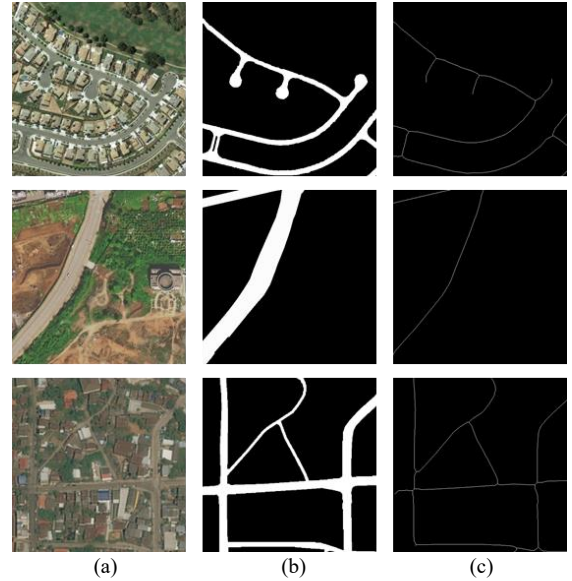

Fig. 4. Examples of the Cheng dataset, Wuhan dataset and DeepGlobe dataset. (a) Image. (b) Road surface ground-truth. (c) Road centerline ground-truth.

### B. Implementation Details

The road label propagation algorithm integrates the buffer-based mask with the graph-based mask to generate the proposal mask. For the buffer inference, $a_1$ was set smaller than the minimum road width and $a_2$ was set close to the maximum road width. For example, we set $a_1 = 6$ m and $a_2 = 18$ m for the Cheng dataset with road widths ranging from 12 m to 18 m. The impact of different buffer widths is further analyzed in Section IV-D. With respect to the graph construction, the SLIC super-pixels were calculated, the approximate number of which was 400 on 512 × 512 image patch, and the compactness was set as 20 to balance the color-space proximity. To find neighbors of the super-pixels, we utilized Delaunay tessellation for simplicity. The color histograms for all the super-pixels were built on the HSV space using 20 bins for the H and S channels; all the bins were concatenated and normalized; and the ranges of H and S were [0, 360] and [0, 1], respectively.

Before training the DBNet, we employed the HED boundary detector which was pretrained on the generic boundaries of BSDS500 [48] to predict HED masks. All the HED masks were seamlessly cropped into 512 × 512 tiles as well. We implemented data augmentation, including image horizontal flip, vertical flip, diagonal flip, color jittering, shifting, and scaling. In terms of the regularized loss, the Gaussian bandwidths for RGB (color domain) and XY (spatial domain) were set as 15 and 100, respectively. The loss weight $\alpha$ of the regularized loss was set to 0.5 in the Equation (3); and the loss weight $\beta$, which balances between the segmentation loss and the boundary loss, was set at 0.7. During the training phase, the Adam optimizer [52] was selected as the network optimizer.

TABLE II
THE EXPERIMENTAL RESULTS OF DIFFERENT WEAKLY SUPERVISED LEARNING METHODS ON THREE ROAD DATASETS,
WHERE THE VALUES IN BOLD ARE WITH THE BEST PERFORMANCE

| Dataset | Method | Precision | Recall | $F_1$ | IoU |
|---|---|---|---|---|---|
| Cheng | ScribbleSup [24] | 0.5274 | **0.9730** | 0.6750 | 0.5190 |
| | BPG [38] | 0.7179 | 0.9085 | 0.7925 | 0.6627 |
| | WSOD [39] | 0.8468 | 0.8852 | 0.8608 | 0.7594 |
| | WeaklyOSM [43] | 0.7798 | 0.9077 | 0.8322 | 0.7170 |
| | ScRoadExtractor (Ours) | **0.9033** | 0.8423 | **0.8657** | **0.7651** |
| Wuhan | ScribbleSup [24] | 0.6086 | **0.7267** | 0.6222 | 0.4740 |
| | BPG [38] | **0.7510** | 0.5804 | 0.6197 | 0.4717 |
| | WSOD [39] | 0.7143 | 0.5329 | 0.5789 | 0.4298 |
| | WeaklyOSM [43] | 0.7509 | 0.6020 | 0.6300 | 0.4805 |
| | ScRoadExtractor (Ours) | 0.6963 | 0.6904 | **0.6580** | **0.5158** |
| DeepGlobe | ScribbleSup [24] | 0.2951 | **0.8813** | 0.4079 | 0.2694 |
| | BPG [38] | 0.6681 | 0.7638 | 0.6624 | 0.5157 |
| | WSOD [39] | 0.6549 | 0.6265 | 0.5899 | 0.4438 |
| | WeaklyOSM [43] | 0.7115 | 0.7378 | 0.6673 | 0.5239 |
| | ScRoadExtractor (Ours) | **0.7954** | 0.7138 | **0.7132** | **0.5782** |

The batch size was fixed as two on the 512 × 512 tiles. The learning rate was initially set at 2e-4, and divided by 5 if the total loss stopped decreasing up to three continuous epochs.

It should be noted that the road label propagation (i.e., buffer inference and graph construction) and HED masks generation were only used for training and not needed for testing. Therefore, in the testing phase, we only applied the DBNet on the test images. The test time augmentation (TTA) was adopted, which included image horizontal flip, vertical flip, and diagonal flip (predicting each image 2 × 2 × 2 = 8 times) also on 512 × 512 tiles. The output probability of each pixel from the sigmoid classifier was translated to binary values with a threshold of 0.5. The road surface extraction results (i.e., the segmentation map derived from the segmentation branch) were evaluated using the four aforementioned metrics.

All the methods were implemented based on PyTorch and the experiments were executed on a single NVIDIA GTX1060 GPU with 6-GB memory.

### C. Experimental Results

We evaluated our proposed ScRoadExtractor on the Cheng dataset, Wuhan dataset, and DeepGlobe dataset and compared its performance with recent scribble-based weakly supervised segmentation methods, including the classic ScribbleSup [24], which adopted an alternative training scheme between proposal generation and network training; Boundary Perception Guidance (BPG) [38], which combined scribbles and rough edge maps for supervision to guide the segmentation network; a weakly-supervised salient object detection (WSOD) method [39]; a method specially for weakly-supervised road segmentation using OSM, named WeaklyOSM [43] in this paper.

As shown in Table II, ScRoadExtractor achieved the best results in both $F_1$ and IoU compared to the other approaches on all the datasets. For ScribbleSup, the alternation between proposal generation and network training happens when training converges; and in this paper, we show its results after three alternations in Table II. Obviously, ScribbleSup performed poorly on these road datasets, and it was computationally expensive due to the alternative training scheme. By contrast, even a single round of training was enough to improve for the end-to-end methods (e.g., BPG, WSOD, WeaklyOSM, ScRoadExtractor). In terms of the Cheng dataset, ScRoadExtractor outperformed BPG by 7.32% in $F_1$ and 10.24% in IoU. Similarly, on the Wuhan dataset and the DeepGlobe dataset, the IoU of ScRoadExtractor was 4.41% and 6.25% higher than BPG, respectively. The BPG had the weakness that the semantics and boundaries interact only at the loss functions without considering the correlation of the features between two sub-networks. ScRoadExtractor performed slightly better than WSOD on the smallest Cheng dataset but significantly better on the Wuhan dataset and DeepGlobe dataset, which indicated WSOD only handled with relatively simple scenarios and lacked of strong generalization ability. Taking the results of WeaklyOSM (the second best) as a baseline, ScRoadExtractor achieved 4.81%, 3.53%, and 5.43% growth in IoU, on the Cheng dataset, Wuhan dataset, and DeepGlobe dataset, respectively.

Figs. 5-7 show some examples of the road segmentation results predicted by different methods on the 512 × 512 tiles selected from these road datasets. Please note that the scribble annotation (b) is not required for testing. It can be seen that the results of ScribbleSup (c) contain many non-road pixels with poor boundary localization. As illustrated in the first rows of Fig. 6 and Fig. 7, BPG (d), WSOD (e) and WeaklyOSM (f) have faced difficulty in correctly identifying the roads shaded by buildings and trees from satellite images, resulting in missing road segments and ambiguous boundaries; but ScRoadExtractor (g) was robust to occlusions and shadows. Furthermore, it can be seen that ScRoadExtractor (g) achieved a segmentation map most similar to the per-pixel annotated road surface ground-truth (h) with better boundary alignment, which demonstrates that ScRoadExtractor can extract the road surface more reliably from remote sensing images.

WeaklyOSM presented in [43] has a problem formulation similar to our study, but ScRoadExtractor differs from it in the following aspects. First, the initial road annotation generation of WeaklyOSM, which had only a buffer-based road width inference, relied on the fixed scribble annotation (i.e., centerline), while other general forms of scribble annotations can be applied by the road label propagation algorithm of ScRoadExtractor. Second, an auxiliary boundary branch was newly employed by ScRoadExtractor to refine and enhance the performance of weakly supervised semantic segmentation. The function of the boundary branch can be seen from Figs. 5-7 that ScRoadExtractor had much better road connectivity through learning from continuous boundaries, and the corresponding ablation study of the next section. To demonstrate the generalization ability of ScRoadExtractor, we introduced the simulated scribbles on the Wuhan dataset that were created by eroding the road surface ground-truth with a cross-shaped kernel of size 7 and offset anchor (3, 6), followed by the skeletonization. Table III shows that our method exceeded WeaklyOSM by 6.44% on $F_1$ and 6.94% on IoU under the same supervision of the simulated scribbles on the Wuhan dataset. In contrast to [43], ScRoadExtractor generalized well on different forms of scribble annotations, ranging from road centerline ground-truth to simulated scribbles, without the limitation of the hard constraint at the early stages.

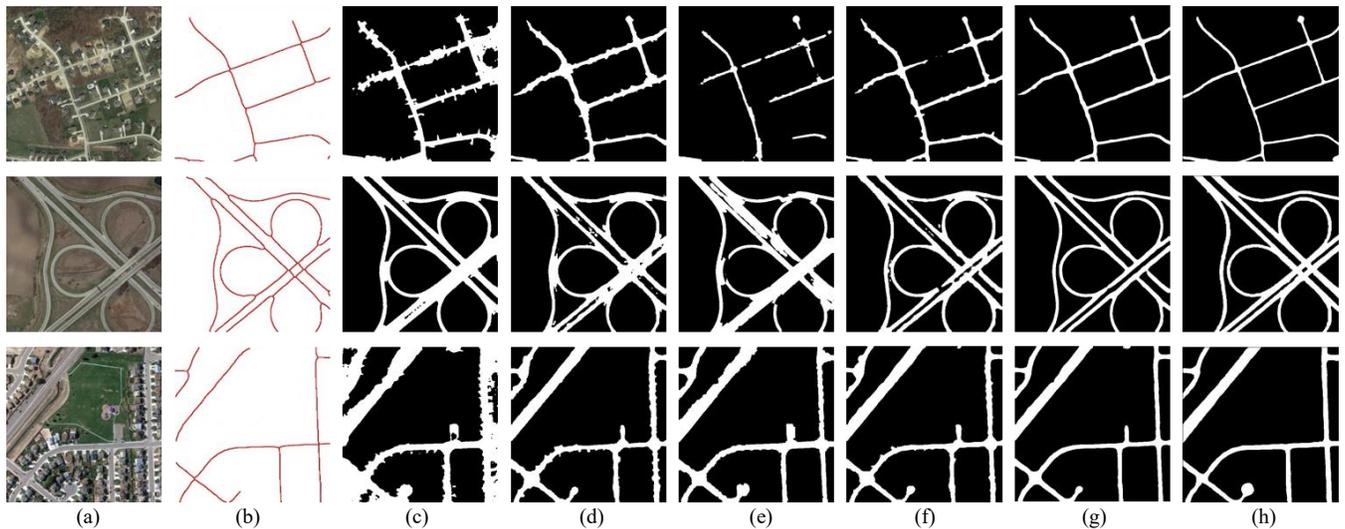

Fig. 5. Qualitative results of road segmentation using different methods on the Cheng dataset. (a) Image. (b) Scribble annotation. (c) ScribbleSup. (d) BPG. (e) WSOD. (f) WeaklyOSM. (g) ScRoadExtractor. (h) Per-pixel annotation (ground truth).

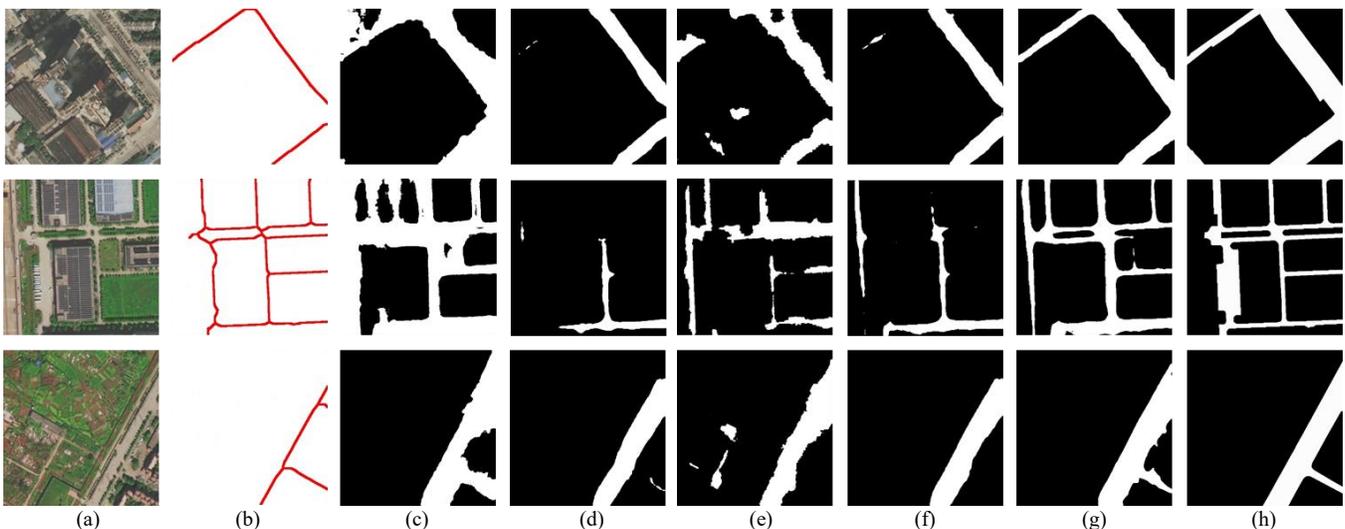

Fig. 6. Qualitative results of road segmentation using different methods on the Wuhan dataset. (a) Image. (b) Scribble annotation. (c) ScribbleSup. (d) BPG. (e) WSOD. (f) WeaklyOSM. (g) ScRoadExtractor. (h) Per-pixel annotation (ground truth).

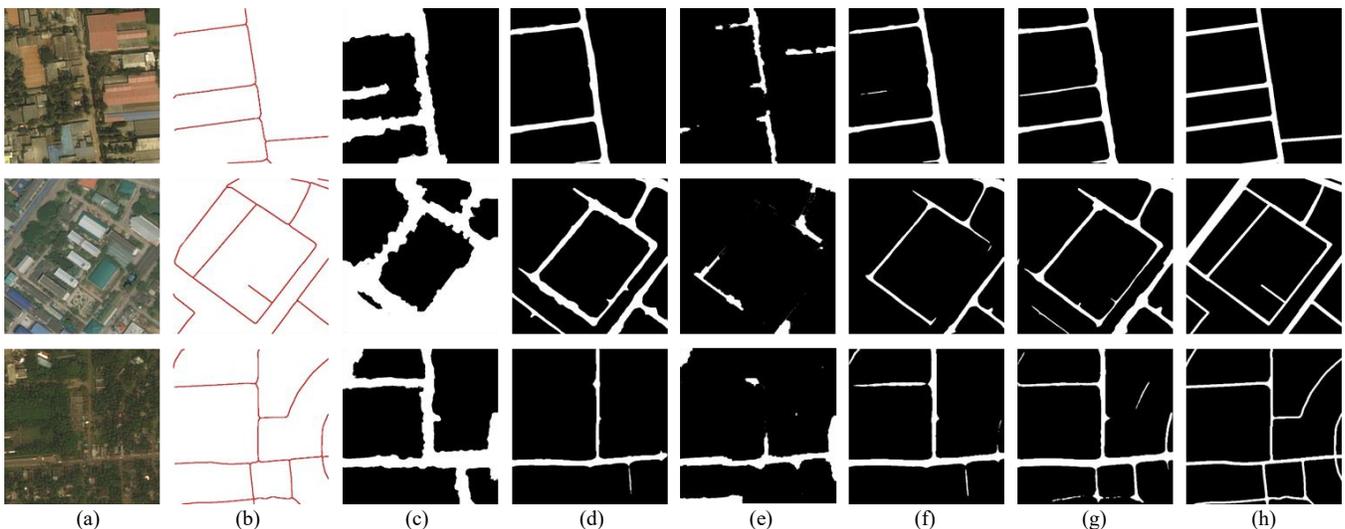

Fig. 7. Qualitative results of road segmentation using different methods on the DeepGlobe dataset. (a) Image. (b) Scribble annotation. (c) ScribbleSup. (d) BPG. (e) WSOD. (f) WeaklyOSM. (g) ScRoadExtractor. (h) Per-pixel annotation (ground truth).

TABLE III
COMPARISON RESULTS BASED ON SIMULATED SCRIBBLES ON
THE WUHAN DATASET

| Method | Precision | Recall | $F_1$ | IoU |
|---|---|---|---|---|
| WeaklyOSM [43] | 0.8661 | 0.4687 | 0.5759 | 0.4307 |
| ScRoadExtractor (Ours) | 0.8318 | 0.5680 | 0.6403 | 0.5001 |

### D. Ablation Study

In this section, the impact of road label propagation algorithm was analyzed by using different buffer widths and different supervision strategies to train DBNet. Additionally, an ablation study was performed to verify the effectiveness of the proposed boundary branch, and we compared it with BRN presented in [38].

First, we evaluated the impact of different buffer widths in proposal mask generation. Parameter $a_1$ defined the scope of pure road pixels. It could be well estimated according to the minimum road width. Here, we focused on the impact of parameter $a_2$, which was a trade-off between buffer inference and graph construction. The results of using different $a_2$ values were shown in Fig. 8. It was observed that the road surface extraction got the best performance when $a_2$ was set at the maximum road width, e.g., 18 m for the Cheng dataset. When $a_2$ was increased or decreased gradually, the performance got worse. If $a_2$ was too large, the proposal mask approached to the empirical buffer-based mask but in fact the road widths varied; if $a_2$ was too small, the proposal mask was closer to the graph-based mask which may contain noise due to the limited capacity of graph cut method. Finally, $a_1$ was set at 6 m for the Cheng dataset, and 2 m for the Wuhan dataset and the DeepGlobe dataset as the latter two cover suburban and rural areas with many narrow roads; $a_2$ was set at 15 m for the Cheng and the DeepGlobe dataset, and 29 m for the Wuhan dataset.

Second, we investigated the effects of weak supervision and full supervision on training the DBNet with different labels. The expanded mask in Table IV indicated directly expanding the centerline with a certain road width (e.g., 10 meters), the buffer-based mask and the graph-based mask were the intermediate products of our proposed road label propagation algorithm (see Fig. 2). As can be seen from Table IV, DBNet trained with our proposed proposal mask obtained the best resu-lts with respect to other weak supervision strategies and achieved acceptable results compared with the per-pixel annotated full mask supervision.

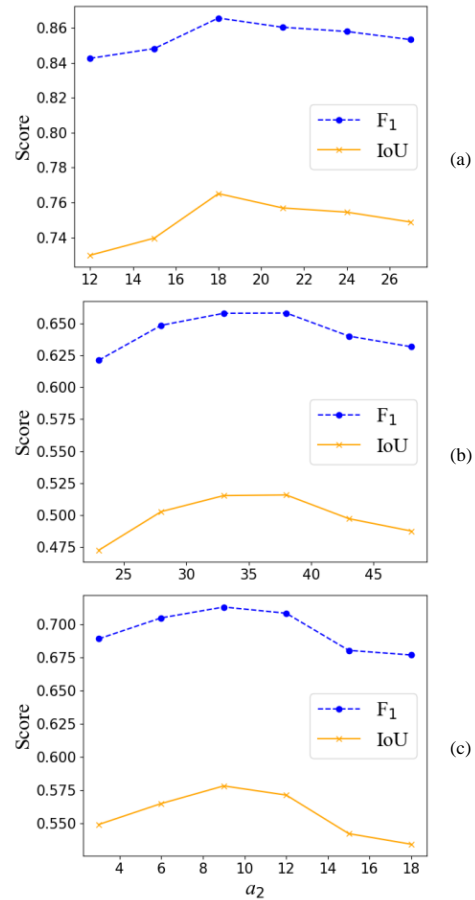

Fig. 8. $F_1$ (blue dotted line) and IoU (yellow solid line) curves for different buffer widths $a_2$ on the Cheng dataset (a), the Wuhan dataset (b) and the DeepGlobe dataset (c).

Third, we explore the impact of the auxiliary boundary branch (short as *Bou*). The comparison results of different branches based on the ResNet-34 backbone were shown in Table V, which were all trained with the same proposal masks. For the Cheng dataset, with the segmentation branch (*Seg*) alone, the model obtained 79.87% in $F_1$. Adding BRN presented in [38], $F_1$ decreased by 3.53%; whereas $F_1$ achieved 4.41% improvement by combining our boundary branch (*Bou*). In terms of the Wuhan dataset, the best results were produced by our proposed DBNet (i.e., w/ *Seg+Bou*), although these network structures performed similarly. Compared with employing *Seg* only and introducing BRN, DBNet improved 3.52% and 1.53% in IoU on the DeepGlobe dataset, respectively. The results demonstrated the generalization ability and effectiveness of our DBNet with the well-designed boundary branch.

There are two critical differences between BRN and our proposed boundary branch. Firstly, only one-shot upsampling layer was used in BRN, while in *Bou* (see Fig. 3) multiple upsampling layers were employed as well as feature sharing with *Seg*. Second, BRN lacked of graph-based regularizations to capture global and local dependencies between known (road and non-road) and unknown pixels. Both of which leaded to it worse performance than ours and even the backbone (w/ Seg) on the Cheng dataset.

TABLE IV
ABLATION STUDY ABOUT DIFFERENT SUPERVISIONS FOR ROAD SEGMENTATION ON THREE ROAD DATASETS

| Supervision | Label | Cheng | | Wuhan | | DeepGlobe | |
|---|---|---|---|---|---|---|---|
| | | $F_1$ | IoU | $F_1$ | IoU | $F_1$ | IoU |
| Weak | graph-based mask | 0.6555 | 0.4972 | 0.6165 | 0.4689 | 0.3980 | 0.2614 |
| | expanded mask | 0.8280 | 0.7089 | 0.5467 | 0.3912 | 0.6203 | 0.4678 |
| | buffer-based mask | 0.8033 | 0.6769 | 0.6325 | 0.4848 | 0.6666 | 0.5224 |
| | proposal mask | 0.8482 | 0.7396 | 0.6484 | 0.5028 | 0.6805 | 0.5422 |
| Full | full mask | 0.9239 | 0.8597 | 0.6956 | 0.5648 | 0.7617 | 0.6408 |

TABLE V
THE IMPACT OF BOUNDARY BRANCH

| Backbone (Encoder) | Branch (Decoder) | Loss | | | Cheng | | Wuhan | | DeepGlobe | |
|---|---|---|---|---|---|---|---|---|---|---|
| | | PBCE | R | $L_{bound}$ | $F_1$ | IoU | $F_1$ | IoU | $F_1$ | IoU |
| ResNet-34 | w/ Seg | √ | | | 0.7987 | 0.6729 | 0.6404 | 0.4944 | 0.6502 | 0.5070 |
| | w/ Seg + BRN | √ | | √ | 0.7634 | 0.6233 | 0.6422 | 0.4948 | 0.6712 | 0.5269 |
| | w/ Seg + Bou | √ | √ | √ | 0.8482 | 0.7396 | 0.6484 | 0.5028 | 0.6805 | 0.5422 |

## V. DISCUSSION

Deep learning has made remarkable achievements in many research subjects, especially vision-based tasks. At the same time, the requirement of huge training datasets is also criticized and rethought. Although the trend of collecting vast amounts of data for deep networks is still on-going, discovering knowledge with less training data (few-shot learning or weakly supervised learning) or without training data (unsupervised learning) has drawn increasingly more attention and begun to form another mainstream. In remote sensing image processing, the lack of high-quality ground-truths has become the norm because a deep learning-based crop-type classifier or road extractor lacks sufficient generalization ability to be applicable to new remote sensing images obtained from dynamic atmosphere conditions without preparing new training data from expensive manual work. How to integrate city-scale or larger scale open-source maps, such as VGI, and the burgeoning weakly supervised learning approach to reduce the demand of training data is a promising area of research as well.

Our proposed ScRoadExtractor is an ideal instance of weakly supervised learning that only utilizes sparse scribble annotations instead of densely annotated ground-truths for road surface segmentation. However, this difference between our method and recent other weakly supervised methods is distinct. The core of ScRoadExtractor is a road label propagation algorithm, which generates the proposal masks from scribbles by aggregating the buffer-based properties of roads and the continuity of similar features in the space and color domains, to mark each pixel as known (i.e., road or non-road) or unknown. Compared to the buffer-based masks (as [43]) and the graph-based masks (as [24]), our proposal masks benefit from a better balance between the utilization of foreground and background information. Based on the proposal masks, we designed a dual-branch encoder-decoder network (DBNet) containing a semantic segmentation branch and a boundary detection branch, which interacted both at the feature level and the output level, while [38] associated them only at the loss functions.

The scribble annotations used in ScRoadExtractor are not restricted to road centerlines from a GIS map or OSM data. Different from [42] and [43], which assume perfect centerlines, more scribble forms can be utilized in ScRoadExtractor; and a commonly-used candidate is GPS traces from vehicles or pedestrians. This kind of scribble can be widely accessed from many open-source databases or websites. Although they are not that accurate, ScRoadExtractor can process them with the combination of buffer-based and graph-based mask generation and boundary alignment. The provided road surface information can in turn aid GPS for better traffic management and navigation.

Our method outperformed state-of-the-art scribble-based weakly supervised segmentation methods as shown in Table IV; nevertheless, it has not fully filled the gap between weak supervision and full supervision as their performance difference is still noticeable. There is a lot of work to be done in future research, which should include the following. First, the graph construction in the road label propagation step can be further formulated as graph representation learning (e.g., via a graph convolution network (GCN) [53]), which can embed the topological information of road networks into the learning-based graph structure. Second, the boundary regularization of road networks will be an important step toward the level of manual delineation. Third, with access to a small number of full (pixel-level) annotations and a large number of weak (scribble) annotations, the proposed method may be able to match the performance of full supervision with semi-supervised learning methods, e.g., an adversarial self-taught learning framework [54] for semi-supervised semantic segmentation.

## VI. CONCLUSION

In this paper, we proposed a scribble-based weakly supervised learning method, called ScRoadExtractor, for road surface segmentation from remote sensing images, which employs an end-to-end training scheme that can achieve good results without the need for alternating optimization. To propagate semantic information from scribble annotations to unlabeled pixels, we introduced a new road label propagation algorithm to generate proposal masks, which integrate the buffer-based masks inferred from the buffer-based strategy and the graph-based masks obtained from the graphs constructed on the super-pixels. In addition, we introduced a dual-branch encoder-decoder network (DBNet), in which we inject the boundary information into semantic segmentation and also a joint loss function that refines both the semantic and boundary predictions. Taking the road centerline as a typical form of scribble annotations in our experiments, we showed that our method was superior to recent related methods and further demonstrated that ScRoadExtractor can be generalized to general forms of scribble annotations.

At present, very few works have explored weakly supervised semantic segmentation for extracting road surfaces from remote sensing images. Our method is a step on a journey that will ultimately bring us closer to automatic road extraction from remote sensing images with very little manual annotating required. We further believe that although ScRoadExtractor was originally designed for road segmentation, we anticipate that it also can be adapted to other segmentation tasks.

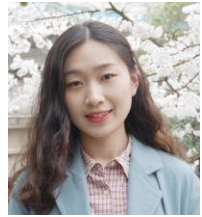

**Yao Wei** received the B.S. degree in geographic information science from China University of Petroleum, Qingdao, China in 2018. She is currently pursuing the M.S. degree in photogrammetry and remote sensing from Wuhan University, Wuhan, China. Her current research interests include machine learning and remote sensing image processing.

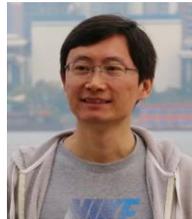

**Shunping Ji** received the Ph.D. degree in photogrammetry and remote sensing from Wuhan University, China in 2007. He is currently a Professor in the School of Remote Sensing and Information Engineering, Wuhan University, China. He has co-authored more than 50 papers. His research interests include photogrammetry, remote sensing image processing, mobile mapping system, and machine learning.